# Belief Updating by Enumerating High-Probability Independence-Based Assignments

**Eugene Santos Jr.**
Dept. of Electrical and Computer Engineering
Air Force Institute of Technology
Wright-Patterson AFB, OH 45433-7765
esantos@afit.af.mil

**Solomon Eyal Shimony**
Dept. of Math. and Computer Science
Ben Gurion University of the Negev
P.O. Box 653, 84105 Beer-Sheva, Israel
shimony@bengus.bitnet

## Abstract

Independence-based (IB) assignments to Bayesian belief networks were originally proposed as abductive explanations. IB assignments assign fewer variables in abductive explanations than do schemes assigning values to all evidentially supported variables. We use IB assignments to approximate marginal probabilities in Bayesian belief networks. Recent work in belief updating for Bayes networks attempts to approximate posterior probabilities by finding a small number of the highest probability complete (or perhaps evidentially supported) assignments. Under certain assumptions, the probability mass in the union of these assignments is sufficient to obtain a good approximation. Such methods are especially useful for highly-connected networks, where the maximum clique size or the cutset size make the standard algorithms intractable.

Since IB assignments contain fewer assigned variables, the probability mass in each assignment is greater than in the respective complete assignment. Thus, fewer IB assignments are sufficient, and a good approximation can be obtained more efficiently. IB assignments can be used for efficiently approximating posterior node probabilities even in cases which do not obey the rather strict skewness assumptions used in previous research. Two algorithms for finding the high probability IB assignments are suggested: one by doing a best-first heuristic search, and another by special-purpose integer linear programming. Experimental results show that this approach is feasible for highly connected belief networks.

## 1 INTRODUCTION

Finding the posterior distribution of variables in a Bayesian belief network is a problem of particular research interest for the probabilistic reasoning community. Although a polynomial-time algorithm for computing the probabilities exists for polytrees [17], the problem was proved to be NP-hard in the general case in [5]. Several exact algorithms exist for computing posterior probabilities: clustering and junction-trees [18, 16], conditioning [6], and term evaluation [19]. These are all exponential-time algorithms in the worst case. Newer algorithms attempt various refinements of these schemes [9].

Several approximation algorithms also exist. In [14] and similar papers, approximation is achieved by stochastically sampling through instantiations to the network variables. In [10], the idea was to use the conditioning method, but to condition only on a small, high probability, subset of the (exponential size) set of possible assignments to the cutset variable.

Recently, approximation algorithms have emerged based on deterministic enumeration of high probability terms or assignments to variables in the network. The probability of each such assignment can be computed quickly: in $O(n)$, or sometimes even (incrementally) in $O(1)$. The probability of a particular instantiation to a variable $v$ (say $v = v_1$) is approximated by simply dividing the probability mass of all assignments which contain $v = v_1$ by the total mass of enumerated assignments. If only assignments compatible with the evidence are enumerated, this approximated the *posterior* probability of $v = v_1$. The approximation improves incrementally as successively more probability mass accumulates.

In [8] incremental operations for probabilistic reasoning were investigated, among them a suggestion for approximating marginal probabilities by enumerating high-probability terms. One interesting point is the skewness result: if a network has a distribution such that every row in the distribution arrays has one entry greater than $\frac{n-1}{n}$, then collecting only $n+1$ assignments, we also have at least $\frac{2}{e}$ of the probability mass. Taking the topology of the network into account, and using term computations, this can presumably be achieved efficiently. However, the skewness assumption as is seems somewhat restrictive. It may hold in some domains, such as circuit fault diagnosing, but certainly not in the typical case, e.g. in randomly generated networks. Slightly relaxing the constraint, say to probability entries greater than $(\frac{n-1}{n})^2$, already requires on the order of $n^2$ assignments to get similar results.

In [21] partial assignments to nodes in the network are created from the root nodes down. The probability of each such assignment is easily computable. Much saving in computational effort is achieved by not bothering about irrelevant nodes (barren nodes), i.e. nodes that are not above some query set node, or nodes that are d-separated from the evidence nodes. Later in that paper, an assumption of *extreme probabilities* is made. This is similar to the skewness assumption above. In fact, in the circuit fault diagnosis experiment in [21], the numbers actually used are well within the bounds of the skewness assumption. The conflict scheme was used later on in that paper in order to narrow the search.

It was already suggested [30, 12] that belief networks frequently have independence structure that is not represented by the topology. Sometimes independence holds given a *particular assignment* to a set of variables $V$, rather than to all possible assignments to $V$. In such cases, the topology is no help in determining independence (e.g. d-separation, which is defined based strictly on topology [20], might not hold), the actual distributions might have to be examined. In [30] the idea of independence-based (IB) assignments was introduced. An assignment is a set of (node, value) pairs, which can also be written as a set of node=value instantiations. An assignment is consistent if each node is assigned at most one value. Two assignments are compatible if their union is consistent. Each assignment denotes a (sample space) event, and we thus use the assignment and the event it denotes as synonymous terms whenever unambiguous. An assignment $\mathcal{A}$ is *subsumed* by assignment $\mathcal{B}$ if $\mathcal{A} \subseteq \mathcal{B}$.

The *IB condition* holds at node $v$ w.r.t. assignment $\mathcal{A}$ if the value assigned to $v$ by $\mathcal{A}$ is independent of all possible assignments to the ancestors of $v$ given $\mathcal{A}_{\text{parents}(v)}$, the assignment made by $\mathcal{A}$ to the immediate predecessors (parents) of $v$.[1] An assignment is IB if the IB condition holds at every $v \in \text{span}(\mathcal{A})$, where span($\mathcal{A}$) is the set of nodes assigned by $\mathcal{A}$. A *hypercube* $\mathcal{H}$ is an assignment to a node $v$ and some of its parents. In this case, we say that $\mathcal{H}$ is *based* on $v$. $\mathcal{H}$ is an IB hypercube if the IB condition holds at $v$ w.r.t. $\mathcal{H}$.

[1]This means that the conditional probability $P(v = d \mid$ assignment to parents of $v)$, where $(d, v) \in \mathcal{A}$, does not change if we condition on any other ancestors of $v$.

In [30], IB assignments were the candidates for relevant explanation. Here, we suggest that computing marginal probabilities (whether prior or posterior), can be done by enumerating high-probability IB assignments, rather than complete assignments. Since IB assignments usually have fewer variables assigned, each IB assignment is expected to hold more probability mass than a respective complete (or even a partial, query and evidence supported) assignment. The probability of an IB assignment is easily computed [30]:

$$P(\mathcal{A}) = \prod_{v \in \text{span}(\mathcal{A})} P(\mathcal{A}_{\{v\}} | \mathcal{A}_{\text{parents}(v)})$$

where $\mathcal{A}_S$ is the assignment $\mathcal{A}$ restricted to the set of nodes $S$. The product terms can each be retrieved in constant time.

One might argue that searching for high-probability assignments for approximating marginal distributions is a bad idea, since coming up with the high-probability assignment is NP-hard [31]. We might have expected that a polynomial time algorithm be sufficient to compute approximations. However, [7] shows that even *approximating* marginal probabilities in belief networks is NP-hard, and thus there *is no* polynomial-time approximation algorithm unless P=NP. Therefore, using this kind of approximation algorithm is a reasonable proposition, provided that for some sub-classes of the problem that are bad for existing algorithms, our approximation algorithm behaves well.

The rest of the paper is organized as follows: section 2 discusses the details of how to approximate posterior probabilities from a set of high-probability IB assignments, and how to modify the IB MAP algorithm of [30] for computing posterior probabilities. Section 3 reviews the reduction of IB MAP computation to linear systems of equations [30], and presents a few improvements that reduce the number of equations. Searching for next-best assignments using linear programming is discussed. Section 4 presents experimental timing results for approximation of posterior probabilities on random networks. We conclude with other related work and an evaluation of the IB MAP methods.

## 2 COMPUTING MARGINALS

The probability of a certain node instantiation, $v = v_1$, is approximated by the probability mass in the IB assignments containing $v = v_1$ divided by the total mass. If we need to find the probability of $v$, then $v$ is a *query* node. Nodes where evidence is introduced are called *evidence* nodes. We also assume that the evidence is conjunctive in nature, i.e. it is an assignment of values to the evidence nodes. We assume that each enumerated IB assignment $\mathcal{A}$ contains *some* assignment to query node $v$, and enforce this condition in the algorithm. Let $I$ be a set of IB enumerated assignments . To approximate the probability of $v = v_i$, we compute:

$$P_a(v = v_i) = \frac{P(\{\mathcal{A}|\mathcal{A} \in I \wedge \{v = v_i\} \in \mathcal{A}\})}{P(\{\mathcal{A}|\mathcal{A} \in I\})}$$

where the probability of a set of assignments is the probability of the event that is the union of all the events standing for all the assignments (not the probability of the union of the assignments). If we are computing the prior probability of $v = v_1$, we can either assume that the denominator is 1 (and not bother about assignments assigning $v$ a value other than $v_1$), or use $1 - P(\{\mathcal{A}|\mathcal{A} \in I\})$ as an error bound. If all IB assignments are disjoint, this is easily computable: simply add the probabilities of the IB assignments in each set to get the probability of the set.

However, since IB assignments are partial, it is possible for the events denoted by two different IB assignments to overlap. For example, let $\{u, v, w\}$ be nodes, each with a domain $\{1, 2, 3\}$. Then $\mathcal{A} = \{u = 1, v = 2\}$ has an overlap with $\mathcal{B} = \{u = 1, w = 1\}$. The overlap $\mathcal{C} = \mathcal{A} \cup \mathcal{B}$ is also an assignment: $\mathcal{C} = \{u = 1, v = 2, w = 3\}^2$. Thus, computing the probability of the union of the IB assignments is non-trivial. We can use the following property of IB assignments:

**Theorem 1** *Let $\mathcal{A}, \mathcal{B}$ be compatible IB assignments. Then $\mathcal{A} \cup \mathcal{B}$ is also an IB assignment.*

Evaluating the probability of a set of IB assignments may require the evaluation of an exponential number of terms. That is due to the equation for implementing the inclusion-exclusion principle of compound probabilities:

$$P(\cup_{1 \leq i \leq m} E_i) = \sum_{k=1}^{m} (-1)^{k+1} \sum_{1 \leq a_1 < \ldots < a_k \leq m} \cap_{i=1}^{k} E_{a_i}$$

where $E_i$ is the ith event.

Fortunately, as we go to higher-order terms, their probability diminishes, and we can ignore them in the computation. That is because low-probability assignments are going to be ignored in the approximation algorithm anyway. How many of the highest probability IB assignments are needed in order to get a good approximation? Obviously, in the worst case the number is exponential in $n$. However, under the skewness assumption [8] (also section 1) the number is small. In fact, it follows directly from the skewness theorem [8] that if the highest (or second highest) probability complete assignment is compatible with $\mathcal{A}_{opt}$ the highest probability IB assignment, and $\mathcal{A}_{opt}$ has at least $\log_2 n$ unassigned nodes, then the 2 highest IB assignments contain most ($\geq \frac{2}{e}$) of the probability mass. It is possible to extend the skewness theorem to include $O(n^k)$ terms, in which case the mass will be at least $\frac{T_k(1)}{e}$, where $T_k(x)$ is the polynomial consisting of the first $k$ terms of Taylor expansion of $e^x$. Thus, under the above conditions, if $\mathcal{A}_{opt}$ has $(k+1)\log_2 n$ unassigned nodes, the highest probability IB assignment will contain at least $\frac{T_k(1)}{e}$ of the probability mass.

[2] Note that for two assignments $\mathcal{A}, \mathcal{B}$, the *union* of $\mathcal{A}$ and $\mathcal{B}$ denotes the event that is the *intersection* of the events denoted by $\mathcal{A}$ and $\mathcal{B}$.

Additionally, all non-supported (redundant) nodes can be dropped from the diagram. A node $v$ is supported by a set of nodes $V$ if it is in $V$ or if $v$ is an ancestor of some node in $V$. A node supported by the evidence nodes is called evidentially supported, and a node supported by a query node is called query supported. We are usually only interested in IB assignments *properly* evidentially supported by some set of evidence nodes. An assignment is properly evidentially supported if all the nodes in the assignment have a directed path of *assigned nodes* to an evidence node. Likewise, an IB assignment is properly query supported if every node in the assignment obeys the above condition w.r.t. query nodes.

Before we start searching for IB assignments, we can drop all evidence nodes that are d-separated from the query nodes, as well as all the nodes that are not either query supported or supported by one of the remaining evidence nodes.

We now present the anytime best-first search algorithm, which is essentially the same as in [30], but with provisions for collecting the probability mass in sets of IB assignments. It keeps a sorted agenda of states, where a state is an assignment, a node last expanded, and a probability estimate:

- Input: a Bayesian belief network $B$, evidence $\mathcal{E}$ (a consistent assignment), a query node $q$.
- Output: successively improved approximations for $P(q = q_i)$, for each value $q_i$ in the domain of node $q$.

1. Preprocessing
   - Initialize IB hypercubes for each node $v \in B$.
   - Sort the nodes of $B$ such that no node appears after any of its ancestors.
2. Initializing: remove redundant nodes, and for each $q_i$ in the domain of $q$ do:
   (a) Set up a result set for $q_i$.
   (b) Push the assignment $\mathcal{E} \cup \{q = q_i\}$ onto the agenda, with a probability estimate of 1.
3. Repeat until empty agenda:
   (a) Pop assignment with highest estimate $\mathcal{A}$ from the agenda, and remove duplicate states (they will all be at the top of the agenda).
   (b) If the assignment is IB, add it to the result set of $q_i$, where $\{q = q_i\} \in \mathcal{A}$, and update the posterior probability approximation.
   (c) Otherwise, expand $\mathcal{A}$ at $v$, the next node, into a set of assignments $\mathcal{S}$, and for each assignment $\mathcal{A}^j \in \mathcal{S}$ do:
      i. Estimate the probability of $\mathcal{A}^j$.

ii. Push $\mathcal{A}^j$ with its probability estimate and last-expanded node $v$ into the agenda.

Expanding a state and the probability estimate is exactly as in [30]: $\mathcal{A}^j = \mathcal{A} \cup \mathcal{H}^j$ is the jth IB hypercube based on $v$ that is maximal w.r.t. subsumption and consistent with $\mathcal{A}$. The probability estimate is the product of hypercube probabilities for all nodes where the IB condition holds. The posterior probability approximation for $q = q_i$ given the evidence is:

$$P_a(q = q_i \mid \mathcal{E}) \quad = \quad \frac{P(\text{result set for } q_i)}{\sum_i P(\text{result set for } q_i)}$$

The preprocessing is independent of the query and evidence sets, and can thus be done once per network. It is also possible to do preprocessing incrementally by moving it into the loop, initializing the hypercubes for a node only when expanded. By using this scheme, it is not even necessary that the belief network be explicitly represented in entirety. Applications which construct belief networks incrementally (such as WIMP [3]) might benefit from not having to generate parts of the network unless needed for abductive conclusions.

It is easy to generalize this algorithm to handle $m > 1$ query nodes, or to compute the probability of a particular joint state of $m$ nodes, or even their joint posterior distribution. This can be done by a somewhat different initialization and estimation steps, which is beyond the scope of this paper.

Experimental results from [30] suggest that at least the highest probability IB assignment (the IB-MAP) can be found in reasonable time for medium-size networks (up to 100 nodes), but that problems start occurring for many instances of larger networks. The idea of using IB assignments to approximate posterior probabilities is independent of the search method. Any algorithm providing the IB assignments in the correct order will do. In the next section, we discuss how the linear programming techniques used in [25, 27, 24, 30] can be used to deliver IB assignments in decreasing order of probability, for posterior probability approximation.

## 3 REDUCTION TO ILP

In [25], [27], [26], and [24], a method of converting the complete MAP problem to an integer linear program (ILP) was shown. In [30] a similar method that converts the problem of finding the IB MAP to a linear inequality system was shown. We begin by reviewing the reduction, which is modified somewhat from [30] in order to decrease the number of equations, and discuss the further changes necessary to make the system find the next-best IB assignments.

The linear system of inequalities has a variable for each maximal IB hypercube. The inequality generation is reviewed below. A belief network is denoted by $B = (G, \mathcal{D})$, where $G$ is the underlying graph and $\mathcal{D}$ the distribution. We usually omit reference to $\mathcal{D}$ and assume that all discussion is with respect to the same arbitrary distribution. For each node $v$ and value in $D_v$ (the domain of $v$), there is a set of $k_{v^d}$ maximal IB hypercubes based on $v$ (where $d \in D_v$). We denote that set by $\mathcal{H}^{v^d}$, and assume some indexing on the set. Member $j$ of $\mathcal{H}^{v^d}$ is denoted $\mathcal{H}_j^{v^d}$, with $k_{v^d} \geq j \geq 1$.

A system of inequalities $L$ is a triple $(V, I, c)$, where $V$ is a set of variables, $I$ is a set of inequalities, and $c$ is an assignment cost function.

**Definition 1** *From the belief network $B$ and the evidence $\mathcal{E}$, we construct a system of inequalities $L = L_{IB}(B, \mathcal{E})$ as follows:*

1. *$V$ is a set of variables $h_i^{v^d}$, indexed by the set of all evidentially supported maximal hypercubes $H_\mathcal{E}$ (the set of hypercubes $H$ such that if $H$ is based on $w$, then $w$ is evidentially supported). Thus, $V = \{h_i^{v^d} | H_i^{v^d} \in H_\mathcal{E}\}$.[3],*
2. *$c(h_i^{v^d}, 1) = -log(P(H_i^{v^d}))$, and $c(h_i^{v^d}, 0) = 0$.*
3. *$I$ is the following collection of inequalities:*
   (a) *For each triple of nodes $(v, x, y)$ s.t. $x \neq y$ and $v \in parents(x) \cap parents(y)$, and for each $d \in D_v$:*
   $$\sum_{(v,d)\in H_i^{x^e}, e\in D_x} h_i^{x^e} + \sum_{(v,d')\in H_j^{y^f}, f\in D_y, d\neq d'} h_j^{y^f} \leq 1 \quad (1)$$
   (b) *For each evidentially supported node $v$ that is not a query node and is not in $span(\mathcal{E})$:*
   $$\sum_{d\in D_v} \sum_{i=1}^{k_{v^d}} h_i^{v^d} \leq 1 \quad (2)$$
   (c) *For each pair of nodes $w$, $v$ such that $v \in parents(w)$, and for each value $d \in D_v$:*
   $$\sum_{i=1}^{k_{v^d}} h_i^{v^d} - \sum_{d'\in D_w \wedge (v,d)\in H_j^{w^{d'}}} h_i^{w^{d'}} \geq 0 \quad (3)$$
   (d) *For each $(v, d) \in \mathcal{E}$:*
   $$\sum_{i=1}^{k_{v^d}} h_i^{v^d} = 1 \quad (4)$$
   (e) *For each query node $q$:*
   $$\sum_{d\in D_q} \sum_{i=1}^{k_{q^d}} h_i^{q^d} = 1 \quad (5)$$

[3] The superscript $v^d$ states that node $v$ is assigned value $d$ by the hypercube (which is based on $v$), and the subscript $i$ states that this is the ith hypercube among the hypercubes based on $v$ that assign the value $d$ to $v$.

The intuition behind these inequalities is as follows: inequalities of type **a** enforce consistency of the solution. Type **b** inequalities enforce selection of at most a single hypercube based on each node. Type **c** inequalities enforce the IB constraint, i.e. at least one hypercube based on $v$ must be selected if $v$ is assigned. Type **d** inequalities introduce the evidence, and type **e** introduces the query nodes. Modifications from [30] include imploding several type **a** equations into one, reducing the number of such equations by roughly a factor quadratic in the number of hypercubes per node. Other modifications are making type **b** and **d** into equalities to make a simpler system, and adding the equations for the previously unsupported query nodes.

Following [25], we define an assignment $s$ for the variables of $L$ as a function from $V$ to $\mathcal{R}$. Furthermore:

1. If the range of $s$ is in $\{0, 1\}$ then $s$ is a 0-1 assignment.
2. If $s$ satisfies all the inequalities of types **a-d** then $s$ is a solution for $L$.
3. If solution $s$ for $L$ is a 0-1 assignment, then it is a 0-1 solution for $L$.

The objective function to optimize is:

$$\Theta_{L_{IB}}(s) = -\sum_{h_i^{v^d}} s(h_i^{v^d}) log(P(H_i^{v^d})) \quad (6)$$

In [30] it was shown that a optimal 0-1 solution to the system of inequalities induces an IB MAP on the original belief network. The minor modifications introduced here, while having a favorable effect on the complexity, encode the same constraint and this do not affect the problem equivalence results of [30].

If the optimal solution of the system happens to be 0-1, we have found the IB MAP. Otherwise, we need to branch: select a variable $h$ which is assigned a non 0,1 value, and create two sets of inequalities (subproblems), one with $h = 1$ and the other with $h = 0$. Each of these now needs to be solved for an optimal 0-1 solution, as in [27]. This branch and bound algorithm may have to solve an exponential number of systems, but in practice that is not the case. Additionally, the subproblems are always smaller in number of equations or number of variables.

To create a subproblem, $h$ is *clamped* to either 0 or 1. The equations can now be further simplified: a variable clamped to 0 can be removed from the system. For a variable clamped to 1, the following reductions take place: Find the type **b** inequality, the type **d** equation (if any), and all the type **a** inequalities, in which $h$ appears. In each such inequality clamp all the other variables to 0 (removing them from the system), and delete the inequality. After doing the above, check to see if any inequality contains only one variable, and if so clamp it accordingly. For example, if a type **d** equation has only one variable, clamp it to 1. Repeat these operations until no more reductions can be made.

| | Min | Max | Avg | Med |
|---|---|---|---|---|
| States | 20736 | 186624 | 84096 | 73728 |

Figure 1: 10 node networks summary. States indicate total number of possible complete assignments in this network.

Once the optimal 0-1 solution is found, we need to add an equation prohibiting that solution, and then to find an optimal solution to the resulting set of equations.

Let $S$ be the set of nodes in the IB assignment $\mathcal{A}$ induced by the optimal 0-1 solution. To update the system, add the following equation:

$$\sum_{v \in S} \sum_{\{H_i^{v^d} | (v,d) \in \mathcal{A}\}} h_i^{v^d} < |S|$$

This equation prevents any solution which induces an assignment $\mathcal{B}$ s.t. the variables in $S$ are assigned the same values as in $\mathcal{A}$. Thus, it is not just a recurrence of $\mathcal{A}$ that is prohibited, but of any assignment $\mathcal{B}$ subsumed by $\mathcal{A}$, in which case we would *also* like to ignore $\mathcal{B}$.

## 4 EXPERIMENTAL RESULTS

As we mentioned earlier, because they are partial assignments, each IB MAP gathers more mass per assignment than the complete MAPs. We studied this mass accumulation for IB MAPs by taking each assignment one at a time in order of probability. By plotting the percentage of mass accumulated versus the number of assignments used, we can get a fair idea of the IB MAP approach's growth rate. In particular, we extracted the top 25 IB assignments per problem instance from 50 randomly generated networks (see [30] for generation method) each having 10 nodes. (We chose 10 nodes since it was still feasible to compute each and every possible assignment in order to get the exact mass.) Figure 1 gives a brief summary of our networks.

Looking at our plot in Figure 2, we can see that mass is accumulated fairly quickly and is contained in a small set of assignments as we expected. After 10 IB MAPs, we have already obtained on average roughly 80% of the total mass (and 60% for the worst diagram instance in the experiment). Note that this result is for unskewed distributions, we expect a far higher accumulation rate for skewed distributions.

With the favorable results for the 10 node cases, we should proceed to the larger network instances. Unfortunately, as we well know, trying a brute force technique of generating all the IB assignments for larger networks is still infeasible. Furthermore, as we mentioned earlier, even the heuristic method for just finding the best IB assignment begins to deteriorate rapidly starting at 100 nodes. Hence, we turn to our linear programming approach. Preliminary results show that our constraints approach can solve for the IB MAP in networks of up to 2000 node. Figure 3 shows the results of our approach on 50 networks each consisting of 2000 nodes. For the most part, we found our solutions relatively quickly. We would like to note though, that our package for solving integer linear programs was crudely constructed by the authors without the additional optimizations such as sparse systems, etc. Furthermore, much of our computational process is naturally parallelizable and should benefit immensely from techniques such as parallel simplex [13] and parallel ILP [1, 2].

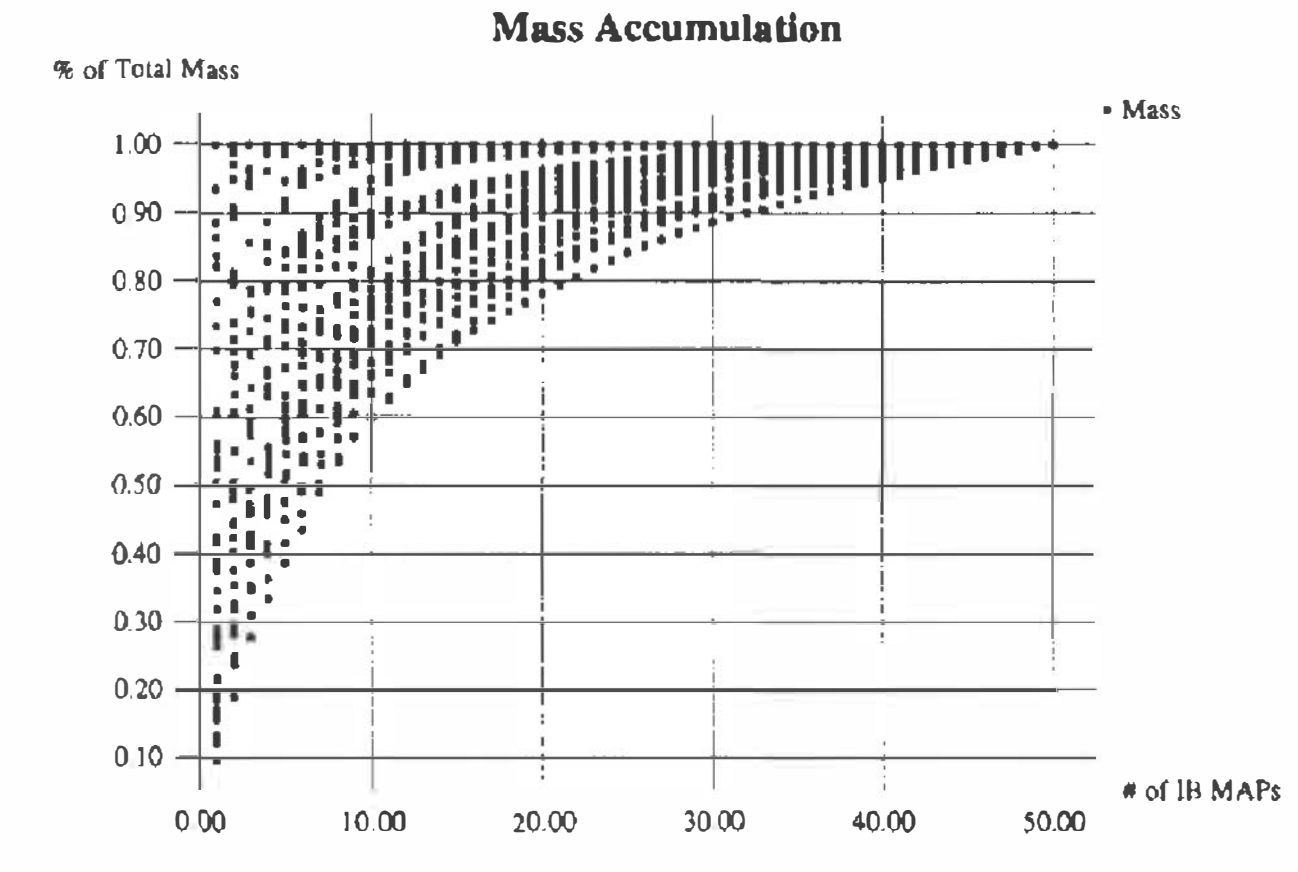

Figure 2: Mass accumulation for 10 node networks.

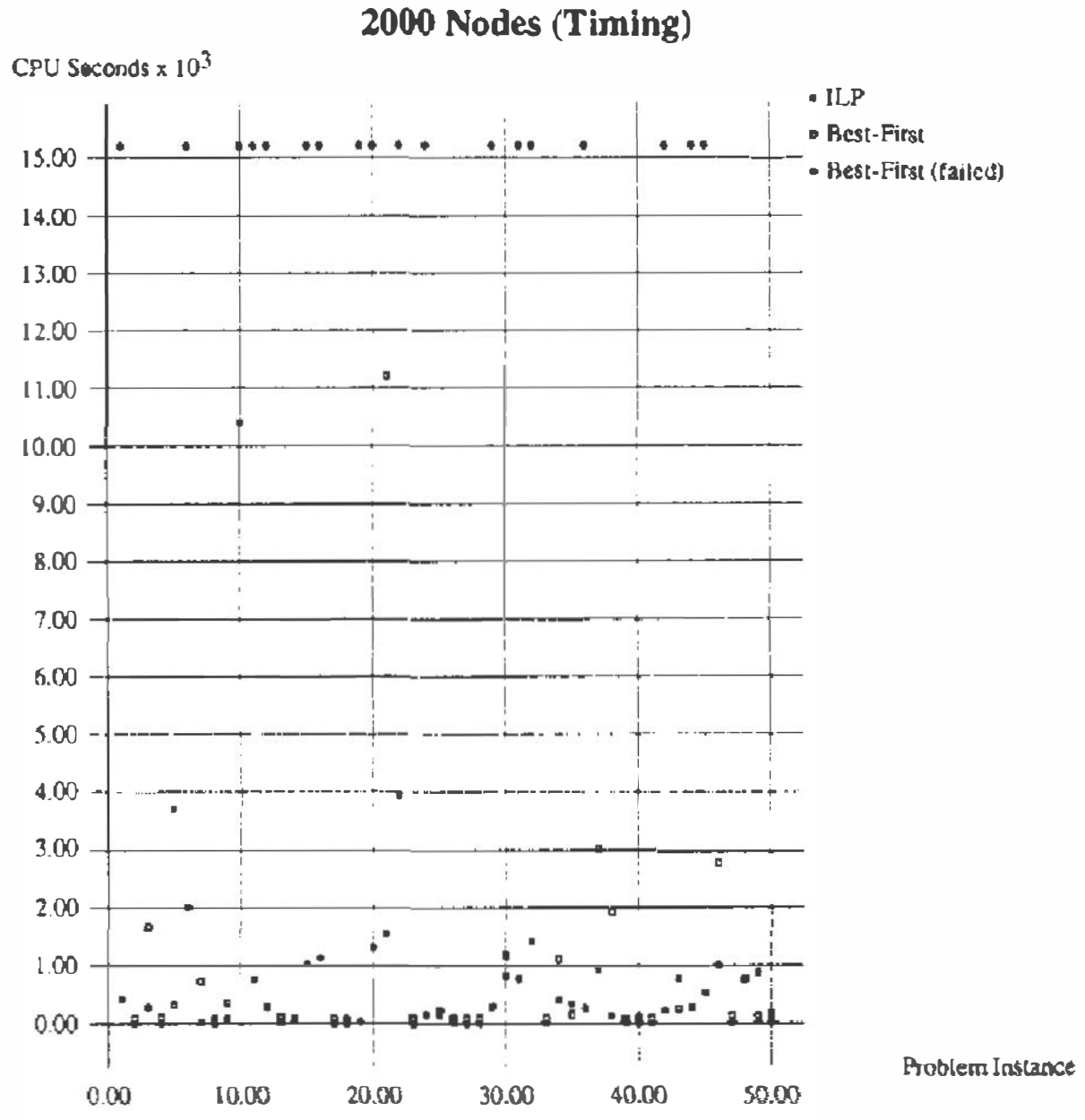

Figure 3: 2000 node networks.

# 5 RELATED WORK

The work on term computation [8] and related papers are extremely relevant to this paper. The skewness assumption made there, or a weaker version of it, also make our method applicable. In a sense, these methods complement each other, and it should be interesting to see whether IB assignments (or at least maximal IB hypercubes) can be incorporated into a term computation scheme.

This paper enumerates high probability IB assignments using a backward search from the evidence. [21] also enumerates high probability assignments, but using a top down (forward) search. Backward constraints are introduced through conflicts. It is clear that the method is efficient for the example domain (circuit fault analysis), but it is less than certain whether other domains would obey the extreme probability assumption that makes this work. If that assumption does not hold, it may turn out that backward search is still better. On the other hand, it may still be possible to take advantage of IB hypercubes even in the forward search approach. It should also be possible to improve performance of the backward search considerably by using a different heuristic than we did. In [4] our heuristic is called "cost so far", and a "cost sharing" heuristic defined there greatly outperforms "cost so far" when applied to proof graphs generated by WIMP [3]. Preliminary attempts to apply cost sharing to finding IB MAPs show a great performance improvement.

The above cited papers [8, 21] as well as this one, are essentially deterministic approximation algorithms. Comparison with stochastic approximation algorithms should also be interesting. Stochastic simulation to approximate marginal probabilities [15] is one such stochastic algorithm. We do not have a ready performance comparison, and the method does not seem immediately applicable to this work.

Other stochastic approximation algorithms find the MAP. For example, in [11] simulated annealing is used. It is not clear, however, how one might use it either to enumerate a number of high-probability assignments or make it search for the IB MAP. A genetic algorithm for finding the MAP [22] makes a more interesting case. The authors in [22] note that the probability mass of the population rises during the search and converges on some value. They do not say whether assignments in the population include duplicates, however, and make no mention of the possibility of approximating marginal probabilities with that population. It seems likely that if the search can be modified to search among IB assignments, then the fact that a whole *population* is used, rather than a single candidate, may provide a ready source of near-optimal IB assignments. Of course, we are not guaranteed to get IB assignments in decreasing order of probability, so slightly different methods would have to be used to approximate the marginal probabilities.

Finally, it should be possible to modify the algorithms presented in this paper to work on GIB assignments and $\delta$-IB assignments, where an even greater probability mass is packed into an assignment [30, 29]. Some theoretical issues will have to be dealt with before we can do that, however.

## 6 SUMMARY

Computing marginal (prior or posterior) probabilities in belief networks is NP-hard. Approximation schemes are thus of interest. Several deterministic approximation schemes enumerate terms, or assignments to sets of variables, of high probability, such that a relatively small number of them contain most of the probability mass. This allows for an anytime approximation algorithm, whereby the approximation improves as a larger number of terms is collected. IB assignments are partial assignments that take advantage of local independencies not represented by the topology of the network, to reduce the number of assigned variables, and hence the probability mass in each assignment.

What remains to be done is to come up with these IB assignments in a decreasing order of probability. This is also a hard problem in general, unfortunately. The factors contributing to complexity, however, are not maximum clique size or loop cutset, but rather the number of hypercubes. Under probability skewness assumptions, the search for high probability IB assignments is typically more efficient, and the resulting approximation (collecting a small number of assignments) is better.

Two algorithms for approximating marginal algorithms are presented: a modification of a best-first search algorithm for finding the IB MAP, and an algorithm based on linear programming. The latter, as expected, proves to be more efficient. We have also experimented on highly connected diagrams where the conditional probabilities are represented as sets of hypercubes (distribution arrays are precluded, since they are exponential in size), and got favorable results in cases where the standard (join-tree or conditioning) algorithms cannot handle in practice.

Future work will attempt to apply the approximation algorithms to cases where the IB condition holds *approximately*, called $\delta$-IB assignments [28]. This should enable representation of *noisy* OR nodes in a linear number of IB hypercubes, where currently this is only possible for perfect or "dirty" OR nodes [30]. Another approach would be to reduce the dimensionality of the conditional tables by using approximation functions [23]. This will directly impact the size of the ILP problem.